\pdfoutput=1

\documentclass[11pt]{article}

\usepackage[final]{acl2023}

\usepackage{times}
\usepackage{latexsym}

\usepackage[T1]{fontenc}

\usepackage[utf8]{inputenc}

\usepackage{microtype}
\usepackage{booktabs}
\usepackage{tabu}
\usepackage{multirow}
\usepackage{graphicx}
\usepackage{hyperref}
\usepackage{subcaption}
\title{Ask an Expert: Leveraging Language Models \\ to Improve Strategic Reasoning in Goal-Oriented Dialogue Models}

\author{
        Qiang Zhang, \textbf{Jason Naradowsky}, \textbf{Yusuke Miyao} \\
        Department of Computer Science \\ The University of Tokyo \\       \{\href{mailto:qiangzhang714@is.s.u-tokyo.ac.jp}{qiangzhang714}, \href{mailto:narad@is.s.u-tokyo.ac.jp}{narad}, \href{mailto:yusuke@is.s.u-tokyo.ac.jp}{yusuke}\}@is.s.u-tokyo.ac.jp
        }

\begin{document}
\maketitle

\begin{abstract}

Existing dialogue models may encounter scenarios which are not well-represented in the training data, and as a result generate responses that are unnatural, inappropriate, or unhelpful.  We propose the ``Ask an Expert'' framework in which the model is trained with access to an ``expert'' which it can consult at each turn.  Advice is solicited via a structured dialogue with the expert, and the model is optimized to selectively utilize (or ignore) it given the context and dialogue history.  In this work the expert takes the form of an LLM.
We evaluate this framework in a mental health support domain, where the structure of the expert conversation is outlined by pre-specified prompts which reflect a reasoning strategy taught to practitioners in the field.  Blenderbot models utilizing ``Ask an Expert'' show quality improvements across all expert sizes, including those with fewer parameters than the dialogue model itself.  Our best model provides a $\sim 10\%$ improvement over baselines, approaching human-level scores on ``engingingness'' and ``helpfulness'' metrics.

\end{abstract}
\section{Introduction}

Dialogue systems based on pre-trained language models (PLMs) can be easily tailored via fine-tuning to exhibit particular characteristics, such as empathy~\cite{roller-etal-2021-recipes} and emotion~\cite{adiwardana2020towards}. 
However, it has been previously observed that such models tend to produce vacuous ``fallback'' responses when presented with unfamiliar situations (e.g., extraneous~\cite{li-etal-2016-diversity,adiwardana2020towards}).
For instance, we observe that fine-tuned BlenderBot~\cite{roller-etal-2021-recipes} models have a propensity to use the response, ``\emph{Do you have any hobbies?}'' as a substitute for furthering the conversation in helpful ways when the situation becomes too complicated. 
For goal-directed dialogues, where the discourse should consistently move towards a desired resolution or effect~\cite{ham-etal-2020-end}, frequent reliance on such fallback responses may result in them performing poorly.

\begin{figure}[t!]
    \centering
    \includegraphics[width=0.48\textwidth]{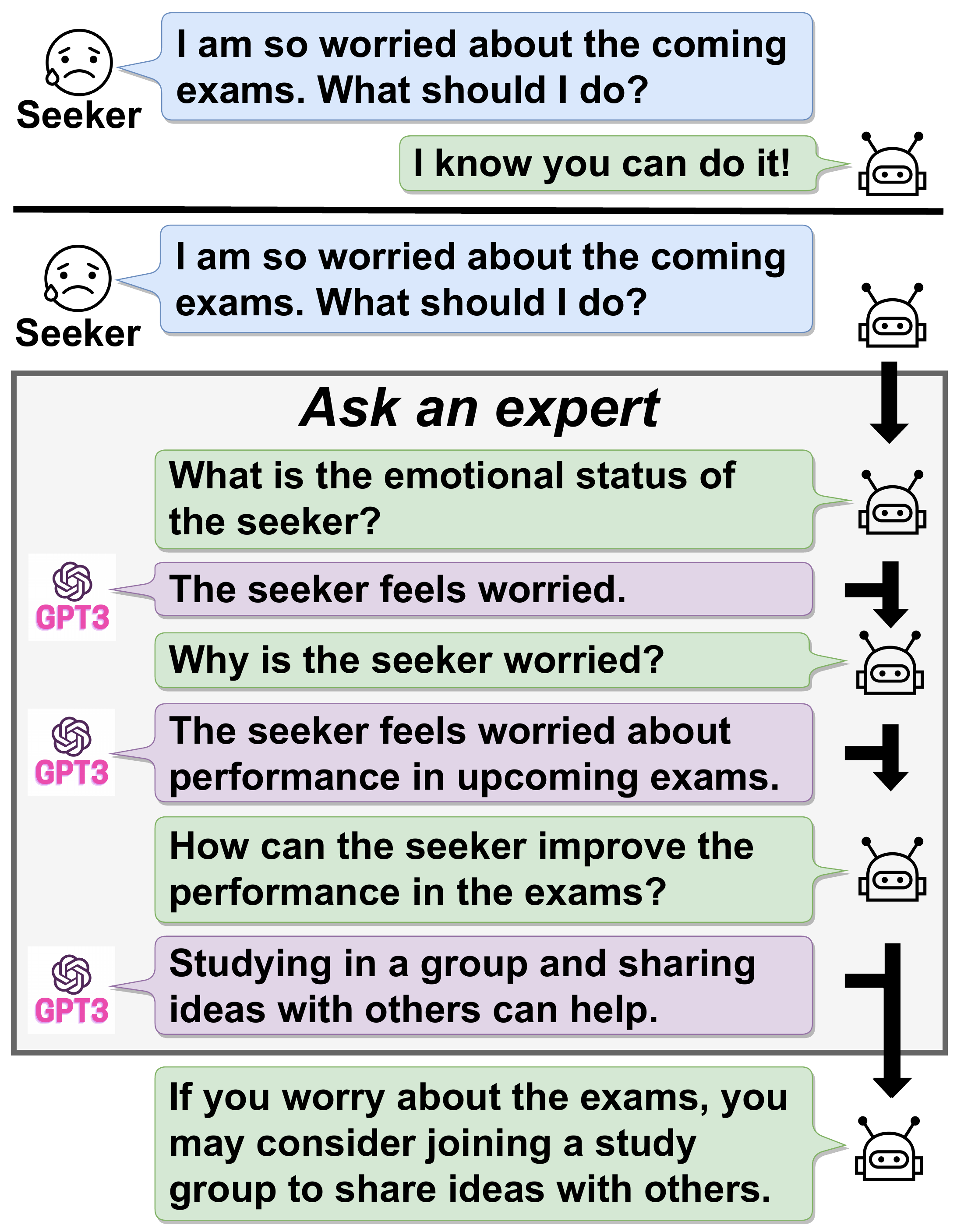}
    \caption{The proposed method of consulting the expert, where the dialogue model interactively obtains advice from the LLM via prompting (e.g. GPT3). Without the aid of expert knowledge and reasoning, dialogue models are less able to generate useful and engaging responses.}
    \label{fig:first_figure}
\end{figure}

We hypothesize that the use of fallback responses may stem from the model being unable to formulate a more suitable reply in the absence of appropriate knowledge of the situation.  In this study, we propose a framework called "Ask an Expert" to enhance dialogue responses through on-the-fly knowledge acquisition. Our approach involves integrating dialogue models with an external ``expert'' by the following tenets: (a) the expert is a large language model (LLM) which is available both during training and inference, (b) the act of soliciting information from the expert itself takes the form of a dialogue, which can span multiple turns in order to identify relevant information and strategies, and (c) the knowledge is integrated into the dialogue model via the context.  Recently many efforts have sought to utilize text as an API to chain together multiple models to perform complex tasks~\cite{shen2023hugginggpt, chase2022langchain}.  Our approach differs in that the model interaction takes place within the optimization loop, and thus allows the dialogue model to learn to selectively choose which advice to incorporate, and when use it.

We apply ``Ask an Expert'' to the domain of mental health support (MHS) systems.  MHS is notable in being one of many domains in which practitioners are formally trained to follow specific discourse strategies~\cite{pudlinski2005doing}.  We incorporate an MHS strategy into the model via a series of hand-crafted prompts, which are designed to shape the expert conversation to reflect the inner monologue of a human expert (Figure~\ref{fig:first_figure}). The resulting conversation is then provided in a structured way as conditioning context to the dialogue model.  

We perform human evaluations on the models following the method of ACUTE-Eval~\cite{li2019acute} to assess the system on six dimensions, including the ability to both have general conversations and provide helpful suggestions.
We find models with reasoning processes significantly outperform the baseline model (without reasoning) in providing constructive suggestions and sharing similar experiences while remaining engaging and empathetic.
Contributions of this work are as follows:

\begin{itemize}
    \item We propose a novel way of formulating knowledge acquisition in dialogue models via a chat-based interaction with a LLM expert, both during training and inference.
    \item We explore several design decisions for structuring the expert reasoning process, and evaluate the effect of different prompts and formats,
    \item We demonstrate that our approach results in dialogues that are deemed more engaging and helpful as evaluated by human judges.
    \item We study the effect of different experts on dialogue quality and present ablation experiments on expert model size.
\end{itemize}

\section{Related Work}

\paragraph{Incorporating Knowledge in Dialogue Models}
Various approaches have been proposed to incorporate external knowledge into dialogue models. Within the scope of deep learning-based models, information may be retrieved from a knowledge base using key-value lookups~\cite{eric-etal-2017-key} or as relation tuples~\cite{young2018augmenting}, or as encoded vectors from knowledge bases~\cite{madotto-etal-2018-mem2seq}. Similar to our work, on-the-fly acquisition of knowledge is possible using the internet as an expert, and integrating search results into the model~\cite{wu-etal-2020-improving-knowledge,komeili-etal-2022-internet}. In addition to relying on external knowledge sources, dialogue models can incorporate knowledge sources, such as pre-trained language models, directly into the decoding process to produce responses grounded in knowledge.~\cite{roller-etal-2021-recipes,xu-etal-2022-beyond,shuster2022blenderbot}. Our approach instead leverage advances in prompt-based text generation and the increasing capacity of LLMs to serve as knowledge bases in order to acquire knowledge as a set of dialogue responses.

\paragraph{LLMs as Source of Expert Knowledge}
Large language models (LLMs) exhibit a remarkable capacity to extract and retain knowledge embedded in the training data. 
Prior studies have demonstrated their ability to extract different forms of general knowledge, including factual knowledge~\cite{petroni-etal-2019-language} and commonsense knowledge~\cite{sap-etal-2020-commonsense}, without requiring fine-tuning. 
Furthermore, LLMs can effectively store and retrieve domain-specific knowledge, such as physical knowledge~\cite{bisk2020piqa} and biomedical knowledge~\cite{yuan-etal-2021-improving}, through knowledge distillation training~\cite{qin-etal-2022-knowledge}.
Prominent models like ChatGPT~\footnote{\url{https://openai.com/blog/chatgpt}} and Bard~\footnote{\url{https://bard.google.com/}} demonstrate impressive proficiency across various natural language processing (NLP) tasks and find practical applications in diverse domains, such as healthcare~\cite{biswas2023role} and finance~\cite{zaremba2023chatgpt}. 
These models not only possess extensive knowledge access but also effectively express this knowledge in natural language, benefiting from instruct-tuning technology~\cite{ouyang2022training} and reinforcement learning from human feedback (RLHF)~\cite{christiano2017deep}.

\paragraph{LLMs for Data Generation and Augmentation}
LLMs can be used to generate additional examples to augment datasets across various NLP tasks and domains, such as text classification task~\cite{wang2021towards}, textual similarity task~\cite{schick-schutze-2021-generating}, and knowledge distillation task~\cite{west-etal-2022-symbolic}.
Unlike previous works, we focus on the data augmentation task for a dialogue dataset in the domain of mental peer support, ESConv~\cite{liu-etal-2021-towards} with additional annotations that come in the form of reasoning support (emotion identification, cause, solution). 

\paragraph{Chatbots for Mental Health}
Given the complexity of providing mental support, rule-based approaches are commonly employed to ensure the generated text adheres to the common behavior of practitioners in the domain.  
For MHS, these guiding rules and principles are agreed upon and proposed by human experts, such as PTSD Checklist~\cite{devault-etal-2013-verbal}, Cognitive Behavioural Therapy (CBT)~\cite{fitzpatrick2017delivering}, Solution-focused Brief Therapy (SFBT)~\cite{fulmer2018using} and mindfulness~\cite{lee2019caring}.   
However, such an approach requires significant efforts to be spent on designing rules and can not handle non-predefined situations. 
Our approach differs in that we reduce the reliance on handcrafting rules by turning to simpler prompt templates, which can then be used together with an LLM to acquire relevant expert knowledge and reasoning for a broad range of different scenarios. 

An alternative is a data-driven approach, wherein deep learning-based dialogue models~\cite{zhang2019dialogpt,adiwardana2020towards,roller-etal-2021-recipes} are trained or fine-tuned on emotion-related datasets such as DailyDialogue~\cite{li-etal-2017-dailydialog}, EmpatheticDialogues~\cite{rashkin-etal-2019-towards}, and EDOS~\cite{welivita-etal-2021-large}.    
Such models are able to produce more empathetic responses, however, possibly due to the lack of explicit strategy, they frequently generate vacuous or unrelated responses.  

\section{Ask an Expert}
The architecture we propose, Ask an Expert, consists of a dialogue model, and a separate expert model.  
In this work the expert is a (presumably larger or specialized) LLM.  
The key distinction between ours and other work which uses additional knowledge acquisition in dialogue systems is that ours takes the form of another dialogue, in which we utilize prompts to guide the expert towards providing important reasoning to guide the dialogue system's response.  The dialogue model is trained to optimize dialogue quality while working together with the expert suggestions, and can therefore learn how best to make use of advice in a context-specific manner.

\subsection{Knowledge Acquisition via Dialogue}
In mental health support (MHS), a seeker (person seeking help) engages in conversation with a supporter (the MHS practitioner) as a way of seeking medical help.  
Like other medical professionals, guidelines and strategies exist for providing mental health support.  
Following the literature, we identify a three-part strategy which involves: (1) identifying the emotional status of the seeker, (2) identifying the reason for that state if undesirable, and (3) providing suggestions that aim to alleviate the underlying cause of the distress~\cite{pudlinski2005doing,tietbohl2022empathic}.  
By designing prompts to collect this information and provide it to the dialogue model, we aim to improve the model's ability to provide useful support and reduce the extent to which it relies on unhelpful fallback responses.

\paragraph{Designing Prompts}
\label{sec:prompt}
\begin{figure}[t]
    \centering
    \includegraphics[width=0.48\textwidth]{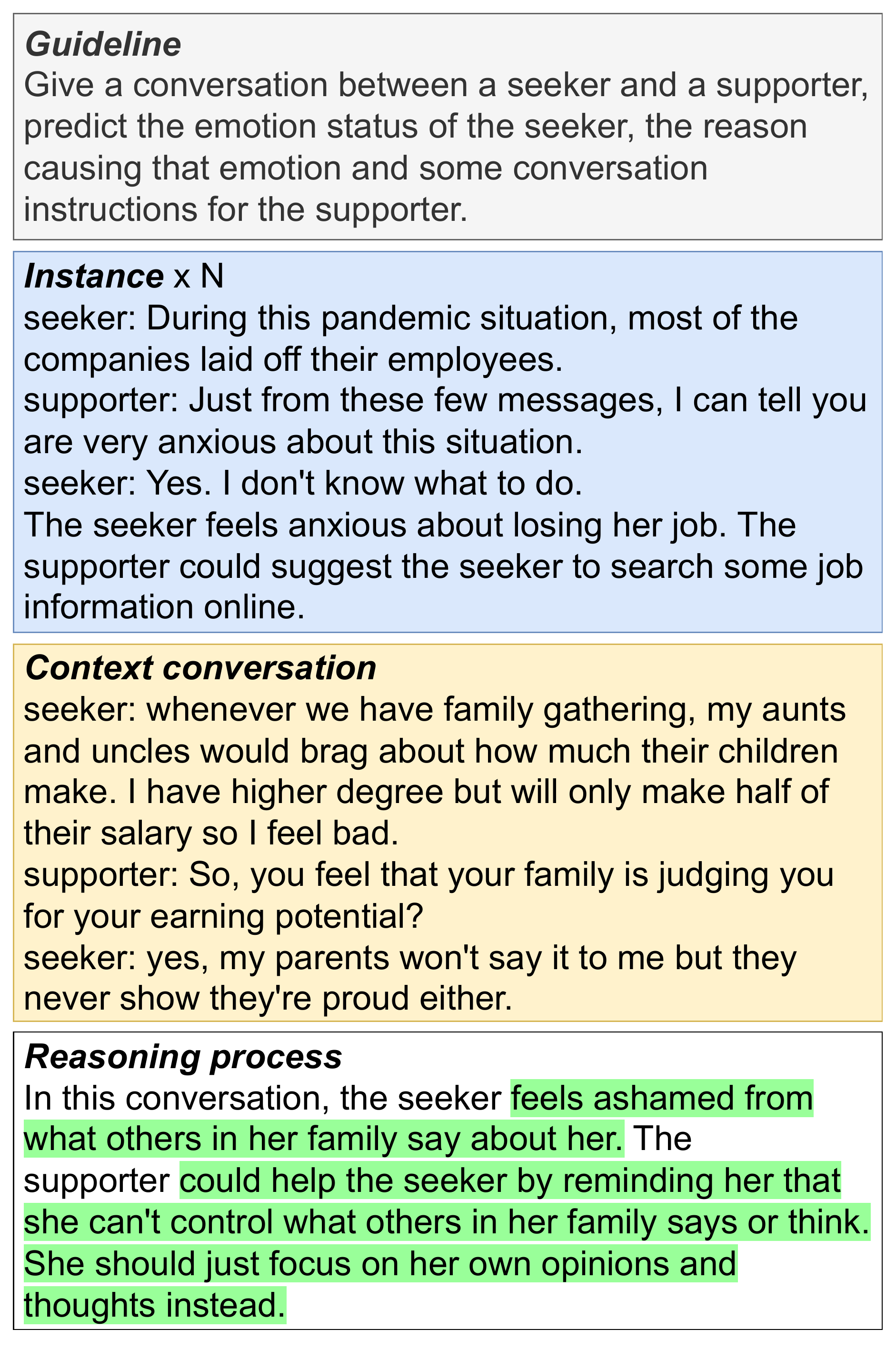}
    \caption{ An example of the dialogue-level prompt used for knowledge acquisition in our setting. The green parts are generated by language models.}
    \label{fig:prompt}
\end{figure}
We compare two different styles of prompts.  
The first, which we refer to ask question-answering (QA), phrases the prompts in the form of questions (e.g., ``\emph{Why does the seeker feel upset with her mother?}'').  The second, which we refer to as text-generation (TG) style echos the masked language modeling objective of LLMs and tasks the model to complete a sentence with missing information (e.g., ``\emph{The seeker feels upset with her mother because...}''). 
Results of our initial experiments comparing the two prompt styles can be found in Appendix~\ref{apx:qa_tg}.  
The remainder of the experiments in this paper use TG-style prompts following the previous works as in ~\citet{schick-schutze-2021-exploiting,mishra-etal-2022-reframing}. 

The second consideration in prompt design is the available length of the prompt.  
We evaluate the Ask an Expert architecture on a variety of base LLMs, ranging in size from GPT to GPT3, meaning that the length of prompts that can fit within the contextual window of the LLMs will vary greatly.  
Hence we designed two different levels of prompt: dialogue-level prompt, in which the instances and context conversation are given as multi-turn dialogue pieces to provide more conversation context, and utterance-level prompt, in which they are reduced to a two-turn dialogue reflecting the current seeker input and the previous supporter's reply.
Figure~\ref{fig:prompt} shows examples of these prompt styles.  Both types of prompts begin with a guideline to describe the task because providing instructions helps LLMs to interpret the task better~\cite{mishra-etal-2022-cross}.
The guideline could also help LLMs to generate the results with the required format as shown in Appendix~\ref{apx:reasoning_samples}.

The context conversation is the history of the preceding dialogue. 
In the utterance-level prompt, several utterances at the beginning of the conversation are trimmed to fit the input length of the LLM.  
The result of this prompted conversation with the expert is a piece of useful information that a human practitioner may very well consider when shaping their responses to the human seeker.  
For instance, a generated reasoning process may be as follows:\\ 

\noindent\textit{``The seeker feels overwhelmed and stressed. He is worried about his upcoming test. The supporter should mention the idea of a study group or a zoom study group. The supporter could also mention Facetime with friends. ''} 

\subsection{Data Collection}

We generate a training set consisting of partial dialogues annotated with the additional reasoning information provided by the expert at each step.  The dialogues are obtained from ESConv~\cite{liu-etal-2021-towards}, a dataset of mental health support dialogues. 
ESConv is especially well-suited for our research because crowdsourcing workers are trained to become supporters when collecting the dataset, and the original annotations on emotion, situation, and strategy can be referred to when designing prompts.

The Ask an Expert architecture is modular, and many models (or humans) could theoretically take the role of the expert.  In this work we wish to assess the importance of model size on reasoning ability and quality of dialogue, and we use the following LLMs as experts: OpenAI GPT (GPT1)~\cite{radford2018improving}, GPT2~\cite{radford2019language}, and GPT3 (ada and davinci)~\cite{brown2020language}.

We balance the data by selecting batches of 8 instances with different combinations of 5 emotion states and 5 problem types (identified from the original annotations in ESConv) with respect to the optimal length of the prompt. In utterance-level prompt situations, the instances are 16 two-turn short conversations. We also empirically adjust the order of instances given the potential influence it could have on the final results~\cite{lu-etal-2022-fantastically}.

We preprocess the conversations in the ESConv dataset, in which speakers can make multiple consecutive utterances, into a turn-based dialogue format by grouping consecutive utterances (if a speaker said, "Why?", and then, "Did anything happen?", they would be combined into a single utterance: "Why? Did anything happen?").  The resulting dataset consists of 9k annotated pairs of seeker-supporter utterances, encompassing 1.5k conversations. We partition the data using a ratio of 70\%/10\%/20\% for training, validation, and testing, respectively.

~\begin{table*}[]
    \centering
    \begin{tabular}{rcccccc}
    \toprule
    \multirow{2}{*}{\textbf{Expert Model}} & \multicolumn{4}{c}{\textbf{Similarity Scores}} & \multicolumn{2}{c}{\textbf{Entailment scores}} \\
    \cmidrule(lr){2-5}
    \cmidrule(lr){6-7}
                  &    BLEU-4   &   ROUGE-L &  BERTScore & BARTScore & RoBERTa & DeBERTa\\
    \midrule
    GPT1         &   0.00      & 0.17      &  86.37     &   - 5.27        & \phantom{0}0.74     & \phantom{0}0.24    \\
    GPT2         &   0.06      & 0.24      &  88.14     &   - 4.41        & \phantom{0}1.23     & \phantom{0}0.74    \\
    ada           &   0.08      & 0.29      &  89.23     &   - 4.04        & \phantom{0}2.81     & \phantom{0}4.06    \\
    davinci       &   \textbf{0.23}      & \textbf{0.46}      &  \textbf{92.03}     &   \textbf{- 3.06}        & \textbf{27.40}               & \textbf{24.44}   \\
    \bottomrule
    \end{tabular}
    \caption{Results of automatic evaluation on the reasoning processes from different PLMs.}
    \label{tab:reasoning_auto}
\end{table*}
~\begin{table*}[ht]
    \centering
    \begin{tabular}{rcccc}
    \toprule
    \multirow{2}{*}{\textbf{Expert Model}} & \multicolumn{3}{c}{\textbf{Voting rates}} \\
    \cmidrule(lr){2-4}
                           &     Emotion Prediction     &   Reason Summarization   &  Suggestion Generation  & Total\\
    \midrule
         GPT1             &    32.23                   &   27.69                 & 21.90                    & 27.27\\
         GPT2             &    44.63                   &   42.15                 & 36.36                    & 41.05\\ 
         ada               &    61.98                   &   57.85                 & 57.85                    & 59.23\\
         davinci           &    \textbf{93.39}          &   \textbf{89.26}        & \textbf{88.17}           & \textbf{90.22}\\
    \bottomrule
    \end{tabular}
    \caption{Human evaluation results three sub-tasks for the information in reasoning processes. Values represent the voting rates of the workers for each sub-task. Total represents overall scores.}
    \label{tab:reasoning_eval}
\end{table*}

\section{Training Dialogue Models}
To evaluate the effect of incorporating our knowledge acquisition procedure into a state-of-the-art dialogue model, we train the following:

\paragraph{Vanilla BlenderBot 2.7B (BB)} The transformer based baseline BlenderBot model fine-tuned on EmpatheticDialogues, ConvAI, WizardofWiki, and BlendedSkillTalks in a multi-task style. We choose this model as the base model because it shows state-of-the-art performance on being empathetic and knowledgable~\cite{smith-etal-2020-put}. 

\paragraph{BlenderBot for Mental Health (BBMH)} A BlenderBot model fine-tuned on the original ESConv dataset, to serve as an in-domain baseline model. BBMH is fine-tuned in a multi-task style on both BlendedSkillTalks and ESConv with equal training weight. This allows BBMH to have a similar conversational ability to BB while having access to mental health-related conversations. 

\paragraph{Blenderbot for Mental Health with Reasoning (BBMHR)} This is a model utilizing the Ask an Expert architecture as applied to mental health support systems, fine-tuned on the reasoning processes that are collected through prompting as described in Section~\ref{sec:prompt}. At training time, seeker utterances and associated reasoning processes that we collected from LLM expert models are concatenated as inputs. 
At inference time, we modify the ParlAI framework to allow communications between the dialogue model and the LLM experts to get ad-hoc reasoning annotations.
Like BBMH, BBMHR is also fine-tuned in a multi-task style on both BlendedSkillTalks and ESConv (with reasoning) for the same purpose.

All models are fine-tuned with ParlAI framework~\cite{miller-etal-2017-parlai} using BlenderBot-BST 2.7B~\cite{roller-etal-2021-recipes} as the initial model~\footnote{The code and data for this work are available at: \href{https://github.com/QZx7/BBMHReasoning/tree/main}{ https://github.com/QZx7/BBMHReasoning/tree/main}}.  Both BBMH and BBMHR are trained on 4 Tesla v100 GPUs for 96 hours. To be noticed, we train multiple BBMHR models with reasoning processes from different LLMs. In the following, BBMHR + \textit{LLMs} denote the dialogue model with reasoning processes from the specific LLM (e.g. BBMHR + \textit{GPT1} denotes the BBMHR model with reasoning processes from GPT1).

\section{Evaluation \& Results}

\subsection{Assessing the Expert Advice}
The first question we aim to answer is: how good is the mental health support advice provided by the LLM experts? 
We perform both automatic evaluation and human evaluation to assess the quality of reasoning processes. 
We randomly select 50 conversations and manually label the conversations (via Mechanical Turk) with reasoning processes.
\begin{table*}[ht]
    \centering
    \begin{tabular}{rccccccc}
    \toprule
    \multirow{2}{*}{\textbf{Model}} & \multicolumn{7}{c}{\textbf{Model Winning Percentages Against Human}} \\
    \cmidrule(lr){2-8}
                   & \small Engagingness  & \small Humanness  & \small Empathy  & \small Specificity  & \small Helpfulness & \small Experience  & \small Total\\
    \midrule
    in-context \textit{davinci} &  - 35.87         &  - 28.89         &  - 24.29         &  - 14.33         &  - 29.65       &  - 24.29 &  -47.30 \\
    \midrule
    BB             &  - 36.78         &  - 22.92         &  - 15.67         &  - 28.91         &  - 30.15         &  - 17.64    & - 42.68\\
    \midrule
    BBMH           &  - 26.07         &  - 21.60         &  - 11.95         &  - 10.53         &  - 22.90         &  - 12.47      & - 30.19    \\
    \midrule
    BBMHR: \\
  \textit{GPT1}   &  - 23.17         &  \phantom{0}- 9.89         &  - 12.51         &  -18.48         &  - 20.07         &  - 10.43      & - 26.20   \\
  \textit{GPT2}   &  - 24.82         &  \phantom{0}- 8.15         &  \phantom{0}- 3.64         &  - 14.02         &  - 19.65         &  \phantom{0}- 9.21      & - 22.33   \\
  \textit{ada}     &  - 24.02         &  \phantom{0}- 7.04         &  \phantom{0}- 7.16         &  - 11.52         &  - 15.59         &  \phantom{0}- 2.48      & - 19.41    \\
  \textit{davinci} &  \textbf{- 12.10}         &  \textbf{\phantom{0}- 1.96}         &  \textbf{\phantom{0}+ 1.26}         &  \textbf{\phantom{0}- 8.60}        &  \textbf{\phantom{0}- 7.09}         &  \textbf{\phantom{0}+ 0.91}      & \textbf{\textbf{- 10.93}}    \\
    \bottomrule
    \end{tabular}
    \caption{Human evaluation results of the winning percentages of different trained dialogue models against human conversations in ESConv. Positive numbers show that the model wins human and negative numbers show that the model loses to human in the comparison.}
    \label{tab:dialogue_eval}
\end{table*}
\paragraph{Automatic Evaluation}
We calculate the similarity and entailment scores between generated reasoning processes and human labels. 
For similarity, we calculate ROUGE~\cite{lin-2004-rouge}, BLEU~\cite{papineni-etal-2002-bleu}, BERTScore~\cite{zhang2019bertscore} and BARTScore~\cite{yuan2021bartscore}. 
Entailment scores are calculated using inferences models, RoBERTa~\cite{zhuang-etal-2021-robustly} and DeBERTa~\cite{he2020deberta} to score the possibilities of the entailment relationship between generated and manual labels by treating it as a textual inference task.

Table~\ref{tab:reasoning_auto} shows the results of automatic evaluation on reasoning processes. 
We can observe clear improvement in both similarity and entailment scores from GPT1 to davinci, where the gap between davinci and other models is especially large.

\paragraph{Human Evaluation}
\label{sec:human_reasoning}
We perform human evaluation to assess the LLMs' ability to generate each piece of information generated in the reasoning processes generation task. 
More specifically, we measure the quality of reasoning processes with three sub-tasks: emotional prediction, reason summarization and suggestions generation. 
Each sub-task is used to assess one piece of information in the reasoning processes. 
Crowdsourcing workers are then asked to vote for each sub-task by answering questions such as ``\textit{Does the annotation contain correct emotion description of the seeker?}''  
We report the voting rates on each sub-task for each expert model used in the prompting phase.   
A complete list of the questions can be found in Appendix~\ref{apx:questions_reasoning}.

Table~\ref{tab:reasoning_eval} shows the results of human evaluation with an average inter-rater agreement of 83.7\%, and we are able to observe similar results as in automatic evaluation. 
Davinci outperforms other models on all three sub-tasks, which shows that davinci may have more knowledge of the reasoning processes.    
Such results hint that the reasoning annotations by consulting LLMs can provide valid reasoning information to be used for dialogue models, especially those generated by LLMs.

\begin{figure*}[t!]
    \centering
    \includegraphics[width=\textwidth]{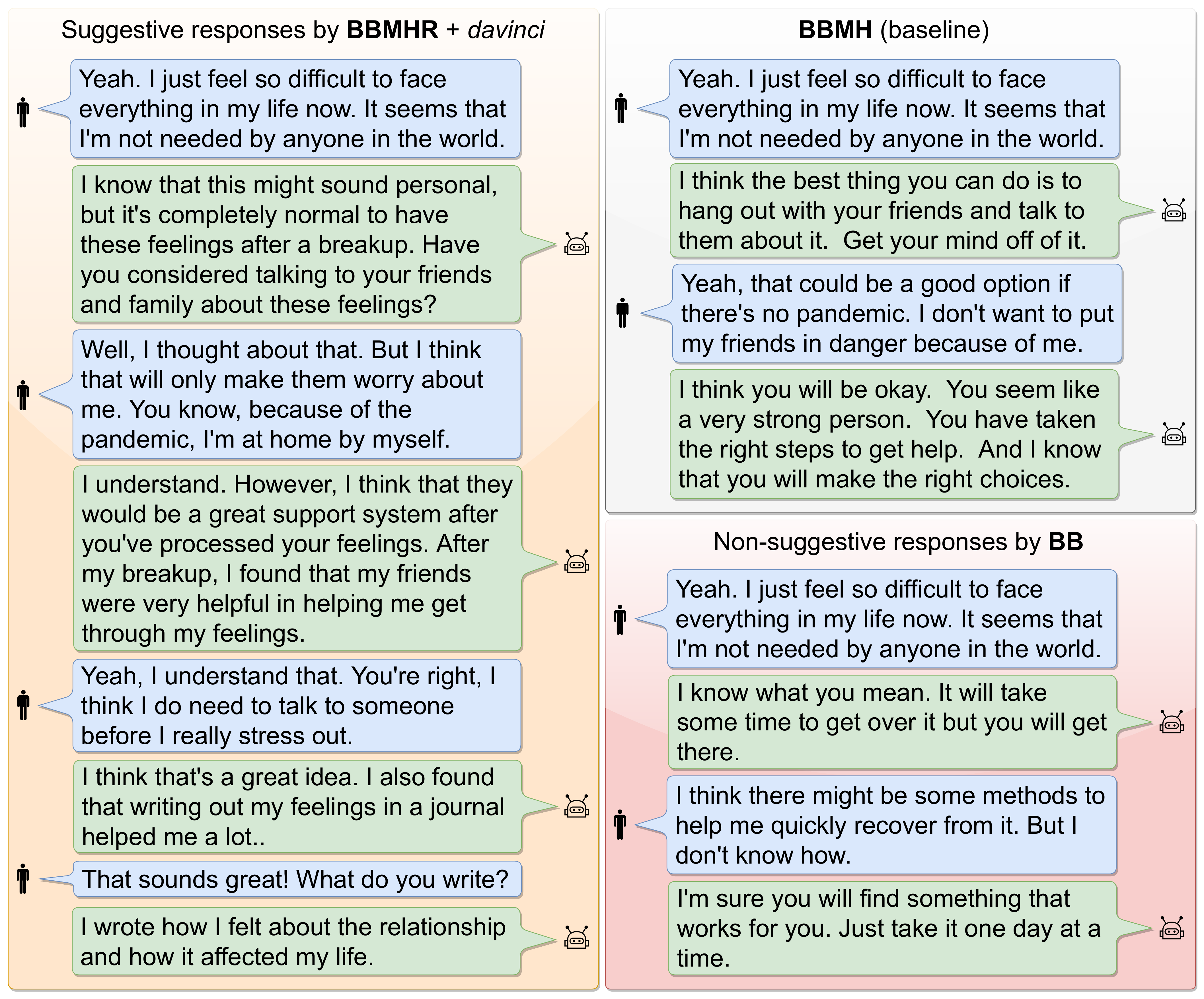}
    \caption{Samples of conversation pieces generated by different models handling the counseling process during the mental support session.}
    \label{fig:sample_counseling}
\end{figure*}
\subsection{Evaluation on Dialogue Models}

We perform the human evaluation on the models following the ACUTE-Eval~\cite{li2019acute} method, in which conversations generated by two different models are collected, and annotators are asked to make binary judgments between two models. 
We set up experiments and compare conversations between humans in ESConv to conversations generated by different models.   
The compared models are divided into three groups: human vs. BB, human vs. BBMH, and human vs. BBMHR.    
For each group, we perform ACUTE-Eval and calculate the win percentages of the models, where positive numbers represent that models win and negative numbers represent that human wins. 
As comparison, we also follow the methods in~\cite{zheng2022augesc} and prompt in-context davinci with the same prompts to generate conversations in the domain of emotional support.


\paragraph{Self-Chats}
We perform self-chats~\cite{jaques-etal-2020-human,bao-etal-2019-know} to collect conversations from models following the experiments in ACUTE-Eval~\cite{li2019acute}. 
Self-chats could reduce the efforts of collecting objective conversations and show high agreements with human-model evaluations~\cite{li2019acute}. 
For each model, we collect 100 conversations across 5 known topics in ESConv, 20 for each topic. 
Initial utterances of the conversations are pre-defined to generate diverse dialogue content for each topic~\cite{bao-etal-2021-plato}. 
The generated conversations are compared against human-human conversations with the same topic in ESConv for evaluation. 

\paragraph{Questionnaire}
Annotators are asked to answer 17 questions across 6 dimensions: engagingness, humanness, empathy, specificity, helpfulness, and experience. 
Engagingness and humanness are used to evaluate the ability to have general and long conversations. Questions for these two dimensions are same as the questions used in~\cite{li2019acute}. 
Empathy represents the model's ability to catch the emotional status and feelings of the seekers. 
Specificity reflects the ability to produce task-specific responses. 
Helpfulness indicates the feasibility of suggestions given by the models. 
Experience is used to measure the ability to share relevant and similar experiences based on the seeker's problems.
We adapted the evaluation method in~\citet{oleary2018suddenly} and crafted questions for the newly added four dimensions based on the components of the ``guided chat tool'', which proved to be more effective in terms of problem-solving.   
A complete list of questions can be found in Appendix~\ref{apx:questions_dialogue}.

\begin{figure}[ht!]
    \centering
    \includegraphics[width=0.48\textwidth]{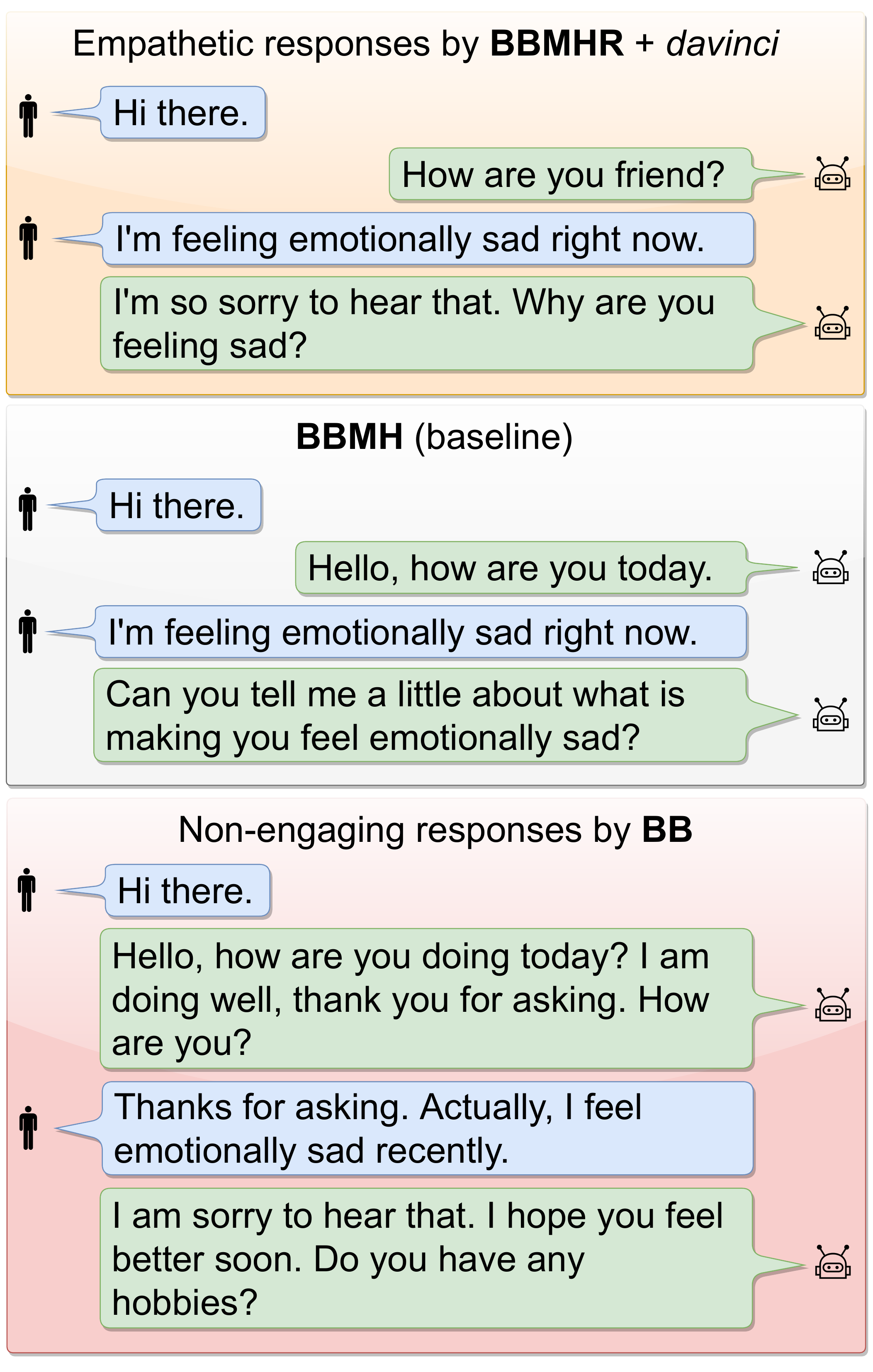}
    \caption{Samples of conversation pieces by different models to initialize the conversation of a mental support session.}
    \label{fig:sample_initializer}
\end{figure}
\paragraph{Results}
Table~\ref{tab:dialogue_eval} shows the results of human evaluation, with an average inter-rater agreement of 80.4\%. Both BBMH and BBMHR outperform vanilla BB in terms of all 6 dimensions, owing to the use of additional in-domain data. 
When assessing the effect of the knowledge acquisition procedure, BBMHR outperforms BBMH in most aspects, especially humanness, helpfulness, and experience, which are the primary criteria that we aim to improve as being especially useful to the goal-oriented aspects of the dialogue model as a mental health support system. 
Additionally, we find a strong correlation with the degree of improvement on these metrics and the size of the model.  Other attributes , such as specificity, do not appear to benefit strongly from additional reasoning information. Among all BBMHR models, BBMHR + davinci achieves the best performance in almost all aspects which also shows that consulting better reasoning models contributes to better responses.

\subsection{Crowdsourcing \& Filtering Details}
The workers are required to be fluent in English in both evaluation tasks of the reasoning processes and dialogue models. 
For reasoning process evaluation, the workers are asked to answer some questions about the content of the conversation to ensure that they clearly understand the context. 
For each question, they also need to provide justifications for their answer to be valid.    
For dialogue model evaluation, while answering the binary selective questions, the workers are asked to write down brief justifications from time to time (Q2, Q5, Q8, Q12, Q14, and Q17) to ensure that they are engaging. 
We perform filtering on the annotations to remove the annotations that are completed in an extremely short time (less than 300 seconds) and with invalid justifications (samples of invalid justifications can be found in Appendix~\ref{apx:crowd_sourcing}). 
The workers are paid an average of 10\$ per hour in line with regional guidelines on ethical compensation.  

\section{Sample Conversations \& Failure Cases}

\paragraph{Sample Conversations}
Figure~\ref{fig:sample_counseling} shows the conversational strategies used by different models when the seeker looks for mental support because of a breakup. 
BBMHR is able to provide suggestive responses based on strategies provided in the reasoning process. 
We also find that BBMHR provides more empathetic and engaging responses when initializing the conversation (In Figure~\ref{fig:sample_initializer}, BB tends to ask non-engaging questions such as ``Do you have any hobbies?''). More samples can be found in Appendix~\ref{apx:sample_conversations}.

\paragraph{Failure Cases}
\begin{figure}
    \centering
    \includegraphics[width=0.48\textwidth]{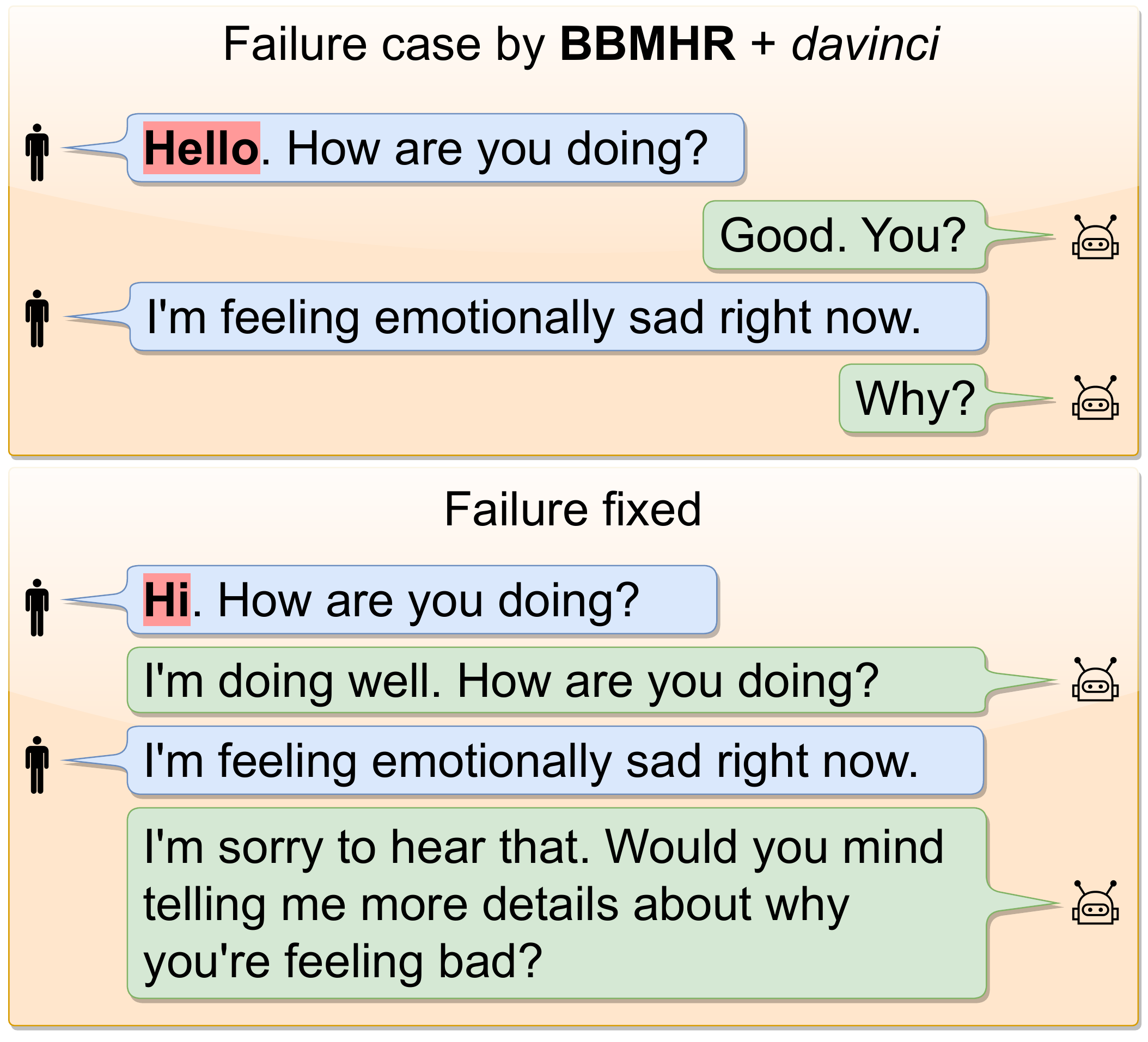}
    \caption{Failure cases by BBMHR + davinci, where the responses of the model are short and non-empathetic. It can be fixed when the opening of the conversation is changed.}
    \label{fig:failure}
\end{figure}
Figure~\ref{fig:failure} shows a failure case where the responses can occasionally be short and not empathetic. 
All models have a tendency to default to such cases at the opening of conversations, when the conversation history is limited and the expert would have difficulty inferring any additional useful details (similar errors are observed in ~\citet{ung-etal-2022-saferdialogues,tyen-etal-2022-towards}).  
Moreover, we observe that the frequency of such failure cases decreases as size of LLM increases, and implies that some of these mistakes may be resolved with better experts. 
For instance, an expert practitioner in this case may be more pro-active in gathering the necessary details to form an analysis.  
By interfacing with the expert purely by text prompts, and collecting the expert advice as text (and inserting it into the dialogue model context window), we allow for the opportunity for the expert model to also help the dialogue model take a more active role in progressing the conversation toward the goal when necessary.

\section{Discussion}

\paragraph{What are the advantages of utilizing LLMs for strategic reasoning?}
Goal-oriented dialogue systems not based upon LLMs often rely on inferring dialogue states to carry out only meaningful conversations,  and thus significantly rely on the definition of the task and an ontology of possible dialogue trajectories~\cite{xie2022converse}.  
This makes the systems brittle and open to catastrophic errors when the dialogue breaks significantly from the categories of the ontology.  
LLMs show similar ontological knowledge and planning ability in many domains, but are more flexible.  
As language models, interfacing with LLM experts is as straightforward as establishing a short goal-oriented conversation, and incorporating their responses into the dialogue model via the model's context is similarly easy.  
In that sense, utilizing LLMs greatly reduces the efforts defining a complicated ontology and dialogue state tracking module by providing necessary reasoning power and knowledge.

\paragraph{Why not use GPT-3 directly for dialogue generation? Is the dialogue model still necessary when there is an expert model?}
Our results (Table~\ref{tab:dialogue_eval}) show that utilizing LLMs as dialogue models directly can lead to worse performance than even baseline dialogue models such as Blenderbot.  We find that in-context davinci performs worse than BB both in terms of generating human-like and empathetic dialogues. One alternative is to fine-tune LLMs specifically for dialogue generation, but this process often requires expensive hardware, time, and training data~\cite{shuster2022blenderbot}.  It is unclear whether fine-tuning even larger models would uncover the heuristic strategies inherent in goal-oriented conversations, which can be easily specified via prompts using an ``Ask an Expert'' architecture.

\paragraph{Deploying Ask an Expert?}
A natural restriction in the Ask an Expert is that it requires the expert to be present at inference time and during deployment. 
If a motivation of Ask an Expert is to allow dialogue models to be deployed on simpler hardware, having a large expert model limits its usefulness in such situations. 
However, recent advancements in technology, such as ChatGPT and Bard, offer API services that facilitate convenient access to expert knowledge. Furthermore, software tools like LangChain efficiently manage prompts, computations, and knowledge, presenting an alternative to local deployment of extensive expert models.

Another scenario that imposes limitations on the adoption of Ask an Expert pertains to certain domains where the system must be deployed locally to uphold privacy concerns, such as mental health systems aiming to safeguard patient data. 
In such instances, relying on external API services becomes less feasible. 
However, it is not always necessary to utilize all the knowledge of large expert models. 
And for specific domain use cases, such as mental health, it is unlikely that the full size of the model is indispensable. 
Given the effectiveness of our approach, in future work we would like to explore the extent to which the expert model can be distilled~\cite{sanh2019distilbert,schick-schutze-2021-just} into models which are able to run locally on consumer-grade hardware.

\section{Conclusion}
In this work we propose the ``Ask an Expert'' framework for building more robust dialogue systems using external knowledge obtained via prompt-based conversations with LLM ``experts''.  The prompts are designed to elicit a step-by-step expert analysis of the current discourse context, intended to mimic the inner monologue of a human professional counselor, and provide it at each turn to the dialogue model.  As the expert consultation process occurs both during training and inference time, the dialogue model itself can learn useful strategies for flexibly incorporating the advice of the expert.  We have shown in both human and automatic evaluations that the addition of such reasoning knowledge results in models which are more suggestive, helpful, and engaging than comparable baseline models which do not consult the expert. 
Our result supports the hypothesis that current dialogue models often fail to implicitly learn effective goal-oriented strategies from dialogue data alone, and provides evidence that combination with other models may help alleviate current shortcomings. 
\section{Limitations and Ethical Considerations}

\paragraph{Limitations}
Our proposed approach relies heavily on LLMs and is subject to the same limitations, namely, known biases in the training data and the ability to hallucinate incorrect information.  Additionally, we perform the research in English only. It is known that for different cultures, the strategies of showing empathy can be very diverse which requires cultural background knowledge and reasoning processes~\cite{atkins2016culture}.  

Pertinent to our intended use-case where models would be deployed locally, LLMs remain computationally intensive even during inference.  Despite demonstrating that even smaller models (such as GPT1 and GPT2) do yield performance enhancements for BBMHR, their performance scales with their parameter size and even small-scale models can require expensive hardware for deployment.  Consequently, it becomes imperative to explore alternative approaches, such as domain-specific lightweight reasoning models, or distilled or low-precision inference models, as viable alternatives to resource-intensive LLMs. 

\paragraph{Ethical Considerations}
Working within the field of mental health support demands additional considerations.  In terms of safety, we acknowledge the limitations of the proposed models and the potential risks associated with directly deploying them to emotionally vulnerable individuals. We do not recommend the deployment of the models presented in this work.  Consequently, we emphasize that the models presented in this study are intended to (at most) function in a human-in-the-loop capacity, serving as an assistant to trained mental health practitioners.

Furthermore, we take into account the possibility of negative impacts that the present research could have on the community. 
Despite our intention to develop models for social good, it is important to acknowledge that the dataset contains content that could be problematic (inputs from seekers, and reasoning processes that could potentially be exploited to generate negative or offensive content).  We release all data collected for this work to help support future work towards improving MHS systems.

\section*{Acknowledgements}
We thank the anonymous reviewers for their helpful
suggestions and feedback.
This work was supported by JSPS KAKENHI Grant Number JP19H05692. 


\bibliography{anthology,custom}
\bibliographystyle{acl_natbib}
\appendix

\newpage
\section{Different Prompts}
\label{apx:qa_tg}
\begin{table*}[b!]
    \centering
    \begin{tabular}{rl}
    \toprule
    Style & Sample \\
    \midrule
         &  \textbf{Context}: \\
         &  \textcolor{blue}{seeker}: I was recently let go from my job due to the covid pandemic, and am now \\
         &        jobless before the holidays. \\
         &  \textcolor{teal}{supporter}: I'm sorry, that's tough anytime but that's gotta be brutal because \\ 
         &  of the year and upcoming holidays.. what kind of work were you doing? \\
         &  \textcolor{blue}{seeker}: It really is, I was in a family owned cafe as a waitress. Due to the rise \\
         &  of the virus they ordered everyone to shut back down. \\
    \cmidrule(lr){1-2}
    QA   &  Q1: How did the seeker feel? \\
         &  A1: The seeker feels tough \textcolor{red}{because of the holiday season}. \\
         &  Q2: Why did the seeker feel that way? \\
         &  A2: The seeker was recently let go from her job and is now jobless. \\
         &  Q3: What could the supporter do? \\
         &  A3: The supporter could tell the seeker that she will help her to find a job. \\
    \cmidrule(lr){1-2}
    TG   &  In this conversation, the seeker feels down because of being jobless. The supporter\\
         & could look for some job openings or tell the seeker to start a small business. \\

    \midrule
         &  \textbf{Context}: \\
         &  \textcolor{blue}{seeker}: I'm very upset. \\
         &  \textcolor{teal}{supporter}: I'm sorry. Would you like to tell me about it?\\
         &  \textcolor{blue}{seeker}: Yes. I invited my friend over to watch my new puppy while I painted my\\
         &  room because my dog has separation anxiety and can't be alone. then she showed\\
         &  up with her new boyfriend. \\
         &  \textcolor{teal}{supporter}: What bothered you about that? \\
         &  \textcolor{blue}{seeker}: Well I don't know him and my stuff was all out in the living room where\\
         &  he was alone for most of the day. I actually didn't suspect anything would\\
         &  happen at first. \\
    \cmidrule(lr){1-2}
    QA   & Q1: how did the seeker feel? \\
         & A1: The seeker feels upset. \\
         & Q2: why did the seeker feel that way? \\
         & A2: The seeker was surprised that her friend brought her new boyfriend over without warning. \\
         & Q3: what could the supporter do? \\
         & A3: The supporter should \textcolor{red}{ask the seeker how she would like to handle the situation}. \\
    \cmidrule(lr){1-2}
    TG   &  In this conversation, the seeker feels upset because her friend brought the new boyfriend \\
         & over to the seeker's house without notifying. The supporter could tell the seeker that it is\\
         & better to talk to the friend about the issue. \\

    \bottomrule
    \end{tabular}
    \caption{Some sample reasoning processes by different styles of the prompt.}
    \label{tab:qa_tg}
\end{table*}
Table~\ref{tab:qa_tg} shows the results by different styles of prompt. We attempted two types of prompt, questions answering (QA) and text generation (TG). In the QA style, we design a series questions asking the information needed for reasoning processes. And for TG style, we prompt PLMs to generate full reasoning processes as a paragraph of natural text.    As shown in the table (text in red color represents the errors), answers by QA style are less accurate and less suggestive than TG style.

\section{Expert Advice samples}
\label{apx:reasoning_samples}
\begin{table*}[h!]
    \centering
    \begin{tabular}{rl}
    \toprule
         Model & Sample \\
        \midrule
                & \textbf{Context}: \\
                & \textcolor{blue}{seeker}: I am ok, I'm having a hard time dealing with the pandemic though. \\
                & ... \\
                & \textcolor{blue}{seeker}: What do you like to do for yourself when you feel bad? \\
                & \textcolor{teal}{supporter}: It happens anyway, but not all the time. when did it start and what do you \\
                &            think that can cause that. \\
                & \textcolor{blue}{seeker}: It started with the pandemic. I think it is because we have to stay home to be \\
                &         safe but other people don't listen and more people are dying because of it. \\
        \midrule
        GPT-1   & the seeker tried to keep her words from being too painful but it was hard and the \\ 
                & source of the pain was more difficult, because her voice was so low. the seeker was \\
                & afraid to talk about this because it was too painful to admit that she had to be truthful. \\
        GPT-2   & feels scared and depressed about losing his job. The supporter could share some \\
                & experience and suggest the seeker to stay positive.\\
        Ada     & feels bad because of the pandemic. The supporter could offer a suggestion. \\
        Davinci & feels bad because of the pandemic. The supporter could tell the seeker to focus on the \\ 
                & things that the seeker can control and try to make a plan for each day. The supporter \\
                & could also suggest the seeker to join an online community to share experiences. \\
        \midrule
                & \textbf{Context}: \\
                & \textcolor{blue}{seeker}: Ok I guess. I do not know how to tell my husband that I am lonely and \\
                & I want out of the marriage. \\
                & ... \\
                & \textcolor{blue}{seeker}: He would go to sleep, and then he thinks I am crazy he says we are no \\
                & kids anymore that need to go on a date we are married. \\
                & \textcolor{teal}{supporter}: Do you think you both might be open to talking to a marriage counselor? \\
                & \textcolor{blue}{seeker}: I want that but I doubt he will talk he tells me he has no problems I have \\ 
                & them, and all it does is I doubt myself. I turned this situation in every \\ 
                & direction and I just am not able to find a solution. \\
        \midrule
        GPT-1   & the seeker felt depressed. she thought about what she had to do now. she did not \\
                & know how to talk to her boyfriend. she could not get him to talk to her but she could \\
                & not stop him from talking to her. she tried to calm down. i 'll talk to him. she thought. \\
        GPT-2   & is worried about losing the job and getting laid off. The supporter could share some \\ 
                & related knowledge and advice.\\
        Ada     & is worried about being alone and not having a husband. The supporter should tell the \\
                & seeker that marriage counseling is available. \\
        Davinci & feels lonely and frustrated. The supporter could suggest the seeker to talk to a marriage \\
                & counselor. \\
                
    \bottomrule
    \end{tabular}
    \caption{Samples of reasoning processes generated by different models.}
    \label{tab:sample}
\end{table*}
Table~\ref{tab:sample} shows the examples of reasoning processes generated by different PLMs used in the experiments.

\section{Questions for Human Evaluation of Expert Advice}
\label{apx:questions_reasoning}
\begin{table*}[]
    \centering
    \begin{tabular}{c}
    \toprule
    Question \\
    \midrule
    Does the reasoning describe correct emotion status of the seeker? \\
    Does the reasoning summarize the seeker's problem correctly? \\
    Does the conversational advice for the supporter make sense? \\
    \bottomrule
    \end{tabular}
    \caption{Questions for human evaluations of the reasoning results.}
    \label{tab:reasoning_questions}
\end{table*}
Table~\ref{tab:reasoning_questions} shows the questions we asked the crowdsourcing workers in the human evaluation of reasoning processes.   Each question is targeting at each piece of information.

\section{Questions for Dialogue Evaluation}
\label{apx:questions_dialogue}
\begin{table*}[h!]
    \centering
    \begin{tabular}{ll}
    \toprule
    Question     &    Choice 1   \\
        \midrule
        \multicolumn{2}{c}{Engagingness} \\
        \midrule
        Which supporter is more engaging to talk to? &  Supporter 1 is more engaging \\
        \textbf{Who would you prefer to talk to for a} & I would prefer to talk to Supporter 1 \\
        \textbf{long conversation?} & \\
        Which supporter do you think is more captivating? & Supporter 1 is more captivating\\ 
         & than Supporter 2\\ 
         
        \midrule
        \multicolumn{2}{c}{Humanness} \\
        \midrule
        Which supporter sounds more human? & Supporter 1 sounds more human \\
        \textbf{If you had to guess that one supporter is human}  & Supporter 1 sounds human \\
        \textbf{and one is a bot, which do you think is human?} & \\
        \textbf{Which supporter sounds more like a real person?} &  Supporter 1 sounds more like a real person \\
        \midrule
        \multicolumn{2}{c}{Empathy} \\
        \midrule
        \textbf{Which supporter understands the feelings} & Supporter 1 understands the feeling better \\
        \textbf{of the seeker better?} & \\
        \textbf{If you had to say one of these supporters} & Supporter 1 understands emotion better \\
        \textbf{understands human emotion better, who would} & \\
        \textbf{ you say is better?}& \\
        Which supporter shows more empathy on the seeker? & Supporter 1 shows more empathy \\
        
        \midrule
        \multicolumn{2}{c}{Specificity} \\
        \midrule
        Which supporter responds more specifically & Supporter 1 talks more relatively \\
        The responses of which supporter are less & Supporter 1's responses are less\\
        out-of-context? & out-of-context\\
        \textbf{Which supporter do you think care more about the} & Supporter 1 cares more about the. \\
        \textbf{seeker's problem?} & seeker's problem\\

        \midrule
        \multicolumn{2}{c}{Helpfulness} \\
        \midrule
        Which supporter gets a stronger urge to help? & Supporter 1 gets a stronger urge to help \\
        \textbf{Which supporter would you prefer to get} & I would prefer to get suggestions \\
        \textbf{suggestions from?} &  from Supporter 1\\
        \textbf{For the suggestions given by the two supporters, }& Supporter 1's suggestion is a better fit \\
        \textbf{which one is a better fit for the seeker?} & than Supporter 2's\\
        
        \midrule
        \multicolumn{2}{c}{Experience} \\
        \midrule
        \textbf{Which supporter shares better similar experience?} & Supporter 1 shares better experience \\
        If you were the seeker, after hearing the experience & Supporter 1's experience would make \\
        of which supporter would you feel better? &  me feel better\\

    \bottomrule
    \end{tabular}
    \caption{Questions for human evaluation of the dialogue models. We design 2-3 questions for each dimensions.}
    \label{tab:dialogue_questions}
\end{table*}
Table~\ref{tab:dialogue_questions} shows the questions we used in the ACUTE-Eval of the dialogue models.  For each dimension, we design 2-3 questions and we calculate the inter-rater agreement for each question to be valid. Bold font indicates that the inter-rater agreements are higher than 85\% and thus are selected for the results calculation.

\clearpage
\section{Interface for Crowdsourcing}
\label{apx:crowd_sourcing}

\begin{figure*}
    \centering
    \includegraphics[width=0.9\textwidth]{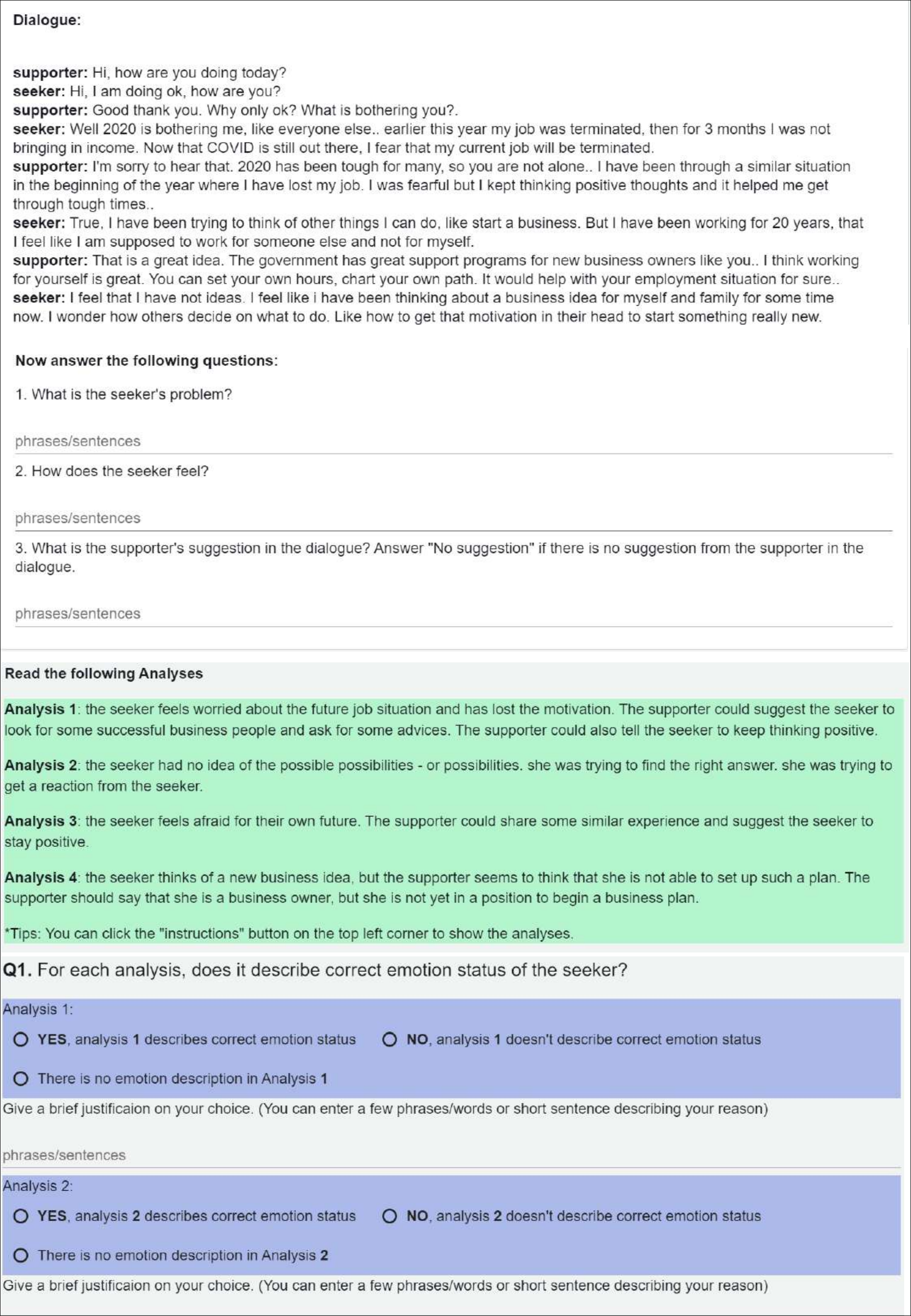}
    \caption{The crowdsourcing interface used to collect evaluation results for the reasoning processes.}
    \label{fig:reasoning_interface}
\end{figure*}

Figure~\ref{fig:reasoning_interface} shows the interface for crowdsourcing that is used in the evaluation of reasoning processes. The crowdsourcing workers are first given the dialogue followed by validation questions asking some details about the conversations. The answers to these questions are then used to filter out invalid questions. Results containing non-sense answers such as ``GOOD, GOOD, GOOD'' are removed from the results. After answering the validation questions, the worker will read through reasoning processes, namely analyses, by different PLMs. The order of the analyses are random for each HIT so that the workers will not capture the pattern for further annotations.   Then for each analysis, the workers are asked to answer the questions in Table~\ref{tab:reasoning_questions}. To be noticed, for each question, the workers will also need to provide a brief justification which will be used as future validation judgement evidence.

\begin{figure*}
    \centering
    \includegraphics[width=0.9\textwidth]{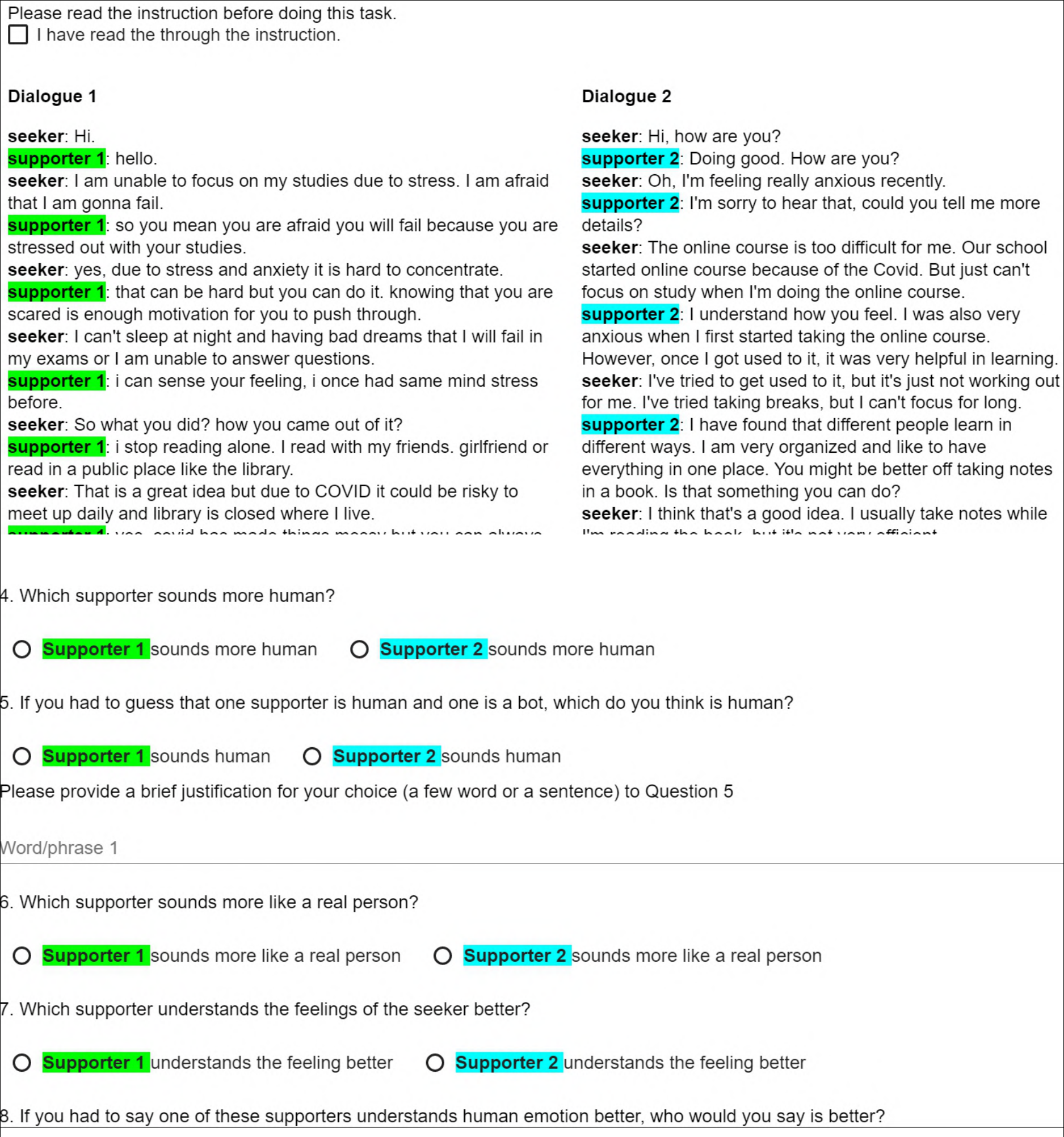}
    \caption{The crowdsourcing interface used for dialogue evaluation.}
    \label{fig:dialogue_interface}
\end{figure*}

Figure~\ref{fig:dialogue_interface} shows the interface we used for ACUTE-Eval of the dialogue models. The workers are first shown two conversations, in which one is directly taken from ESConv, namely human-human and one is generated by the self-chats of the model. The order of the conversations are randomly selected for each HIT. After reading the two conversations, the workers are then asked to answer the questions listed in Table~\ref{tab:dialogue_questions}.  From time to time, we ask the workers to provide brief justifications for their choice and such justifications will be used to filter out invalid results.

\newpage
\section{Responses that apply 'online' strategy in ESConv}
\label{apx:responses_online}
\begin{table*}[]
    \centering
    \begin{tabular}{l|l}
    \toprule
         \textbf{Seeker} & \textbf{Supporter} \\
    \midrule
    \multicolumn{2}{l}{\textit{Ongoing depression on pandemic}} \\
    \cmidrule(lr){1-2}
         Yes, I pay musical instruments but do to COVID & Could you perhaps set up Zoom meetings \\
         could not play with the band. &  where you could play together \textcolor{blue}{online}?\\
    \cmidrule(lr){1-2}
         Hmm what specific hobbies would you  & Whichever you enjoy.. pick one. There are a \\
         recommend? & lots of \textcolor{blue}{online} resources you cloud use. \\
    \cmidrule(lr){1-2}
         Do you have any suggestions? & You can play \textcolor{blue}{online} games with your friends. \\
    \cmidrule(lr){1-2}
         That actually sounds like a good idea. I hope  & If you are not comfortable going out due to \\
         the shelter near me will take volunteers with & COVID, you could involve some activities  \\
          COVID and all. & \textcolor{blue}{online} promoting dog adaption and create \\
         &  awareness \textcolor{blue}{online} and through social media... \\
    \cmidrule(lr){1-2}
         All I have to do is think about how alone I am. & Do you have any friends or people you can set\\
         &  up an \textcolor{blue}{online} zoom call with? \\
    \cmidrule(lr){1-2}
         I have tried to use zoom and facetime but video & There are \textcolor{blue}{online} resources to have some fun \\
         chat gives me anxiety. & with friends too--many blogs suggest hosting \\
         & a group game night or a shared movie night. \\
    \midrule
    \multicolumn{2}{l}{\textit{Job crisis}} \\
    \cmidrule(lr){1-2}
        Hmm that seems like a good idea, to find video to & well for me i just searched for motivational \\
        help uplift me. Do you recommend anything? & speaker or top 10 \textcolor{blue}{online}?work from home jobs. \\
    \cmidrule(lr){1-2}    
        yes It is my main concern. & Have you consulted with a job center, a life \\
        & coach, or any other resource such as \textcolor{blue}{online} \\
        & websites? These may be useful.\\
    \cmidrule(lr){1-2}    
        Yes , I also dont want them to have to support me & with keeping your family in mind while trying \\
        and my family either . & to find a job have you considered looking for \\
        &  an \textcolor{blue}{online} job? Just from chatting with you I can \\
        & tell how much it stresses you out. \\
    \cmidrule(lr){1-2}    
        I would be open to seeking other employment & Luckily, there are many platforms \textcolor{blue}{online} \\ 
        online;work from home on the computer.  & that allow you to work from home. I know \\ 
        any suggestions? & of several that allow you to do side \"gigs\". \\ 
        & Perhaps you can search and find a few of these. \\
        & I, myself have had success doing these..\\
    \cmidrule(lr){1-2}    
        I found it really difficult finding a job right now & Have you tried searching a job from some \\
        because of the pandemic. & \textcolor{blue}{online} job-hunting platforms? \\
    
    \bottomrule
    \end{tabular}
    \caption{Some sample responses under the topic of ongoing depression and job crisis because of COVID pandemic in ESConv. 75\% percent of the responses are replying about using online resources (online meeting, online gaming, online party, etc.)}
    \label{tab:ongoing_responses}
\end{table*}
The responses tend not to follow the reasoning from PLMs when same strategies are frequently repeated in the training data of ESConve for the conversation with same context. From the collected conversations, we are able to find that in most cases, BBMHR will follow the suggestions in annotations.  And for all the cases where BBMHR doesn't follow the suggestions, they follow frequently repeated strategies applied in the training data of ESConv. For instance, one case where BBMHR tends to not follow the reasoning annotations is in the topic of ongoing depression. When the seeker inputs like ``I feel really depressed because of the pandemic. '', BBMHR tends to produce a response like ``Have you tried hanging out with your friends online?'' even the reasoning annotation is like ``The supporter could suggest the seeker to go out and take a break.''  And in ESConv, we are able to find that more than 75\% of conversations with the topic of ongoing depression have applied similar responses.  Such ignorance of reasoning annotations also happens in the context of job crisis where ``searching for online information'' is a repeated strategy.   However, the ignorance of reasoning annotations do not appear for other topics that do not share a frequently repeated strategy.

Table~\ref{tab:ongoing_responses} shows examples of frequently repeated answers and strategies in the ESConv dataset that can affect the responses. When the BBMHR models take such context as input, they tend to ignore the reasoning processes from PLMs and follow the strategies stated in the dataset.

\clearpage
\section{Sample Conversations from Different Models}

\begin{figure*}[t!]
    \centering
    \includegraphics[width=\textwidth]{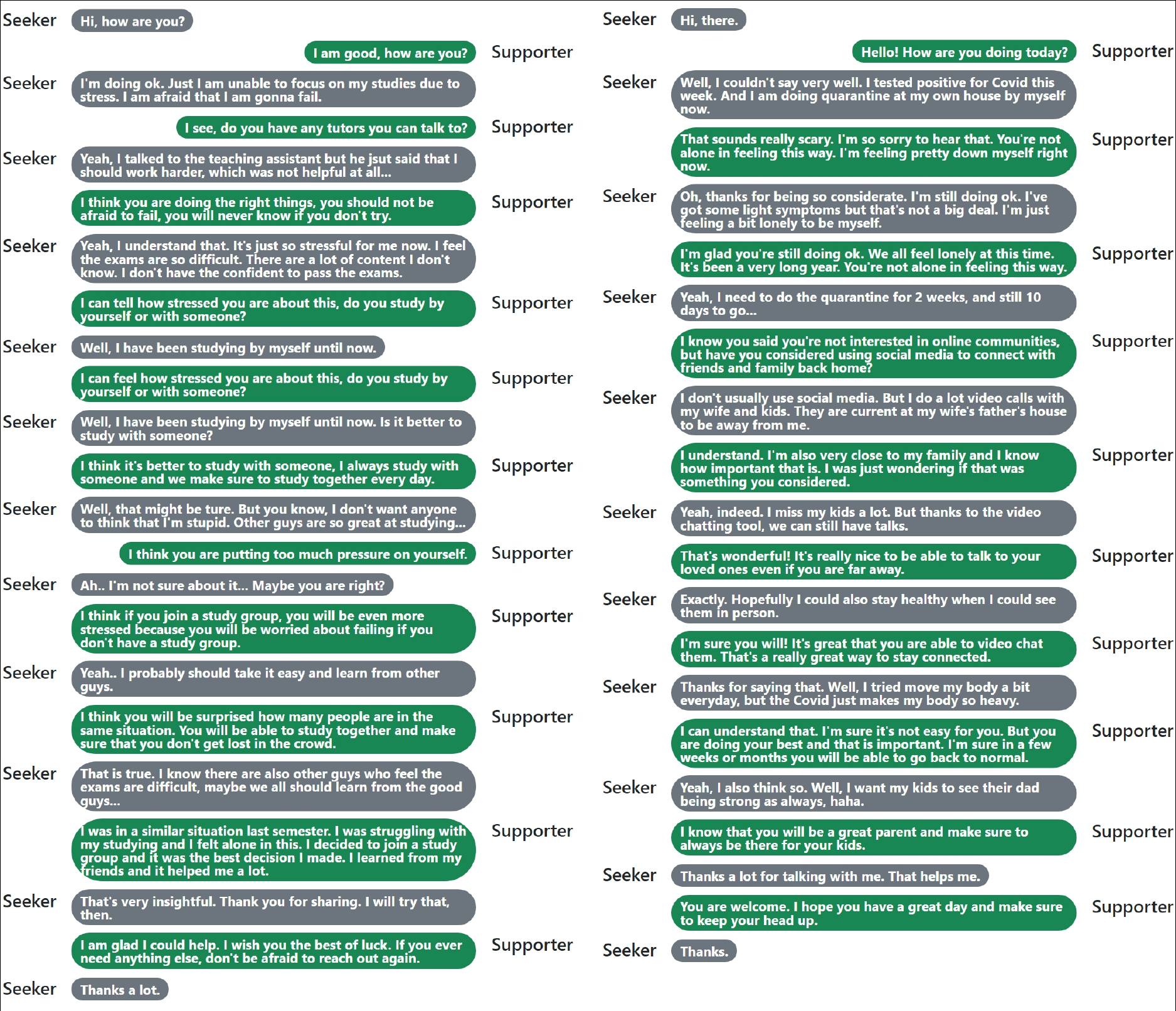}
    \caption{Sample conversations by BBMHR + \textit{davinci} on the topics of ``academic'' and ``ongoing depression''.}
    \label{fig:sample_davinci}
\end{figure*}

\begin{figure*}
    \centering
         \includegraphics[width=\textwidth]{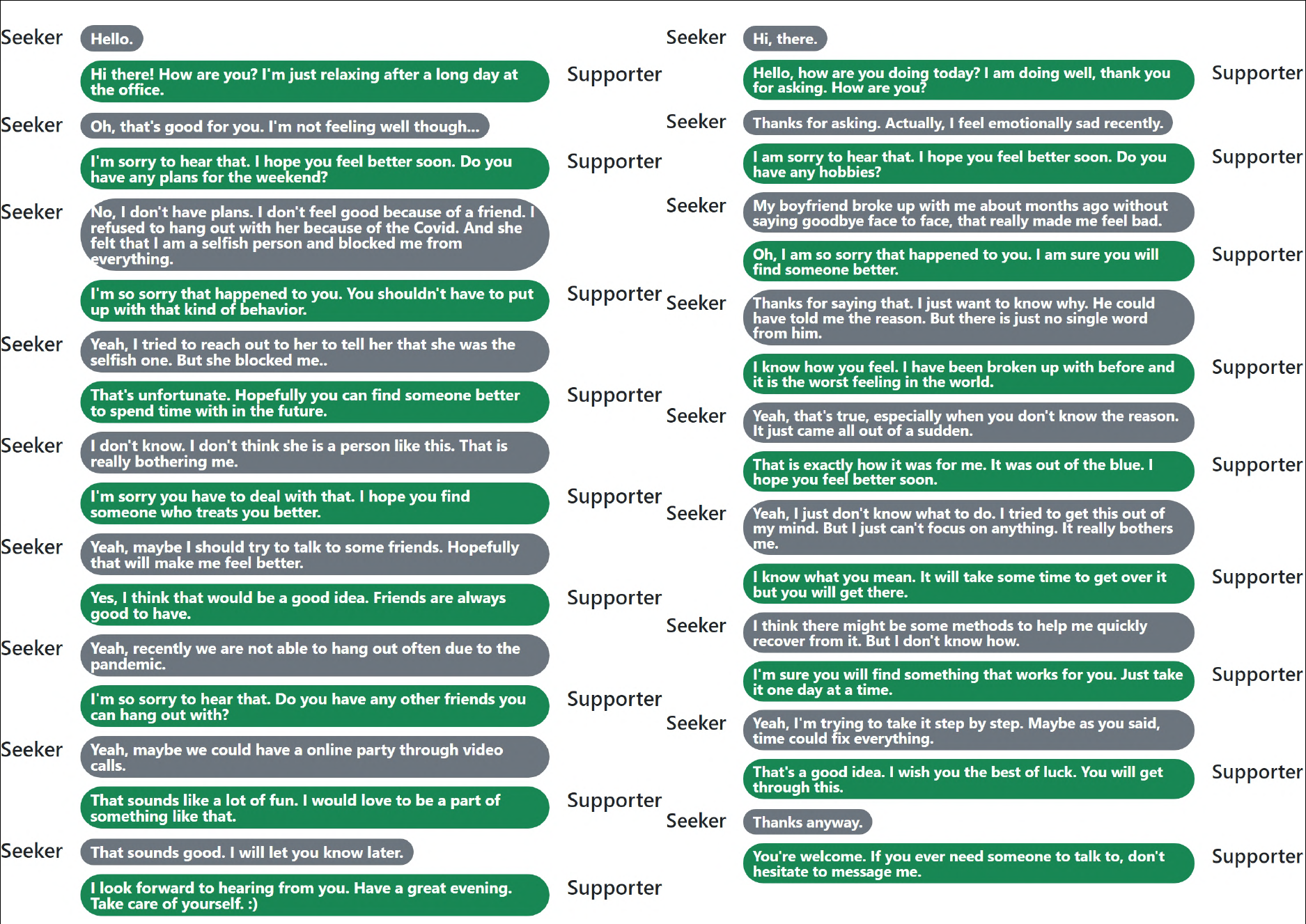}
        \caption{Sample conversations by BB on the topics of ``problems with friends'' and ``break up with partner''.}
    \label{fig:sample_bb}
\end{figure*}

Figure~\ref{fig:sample_davinci} \~~\ref{fig:sample_bb} show sample conversations generated by BBMHR, BBMH and BB models on various topics. We are able to observe generally more specific and suggestive responses from BBMHR models.
\label{apx:sample_conversations}

\begin{figure*}
    \centering
        \includegraphics[width=\textwidth]{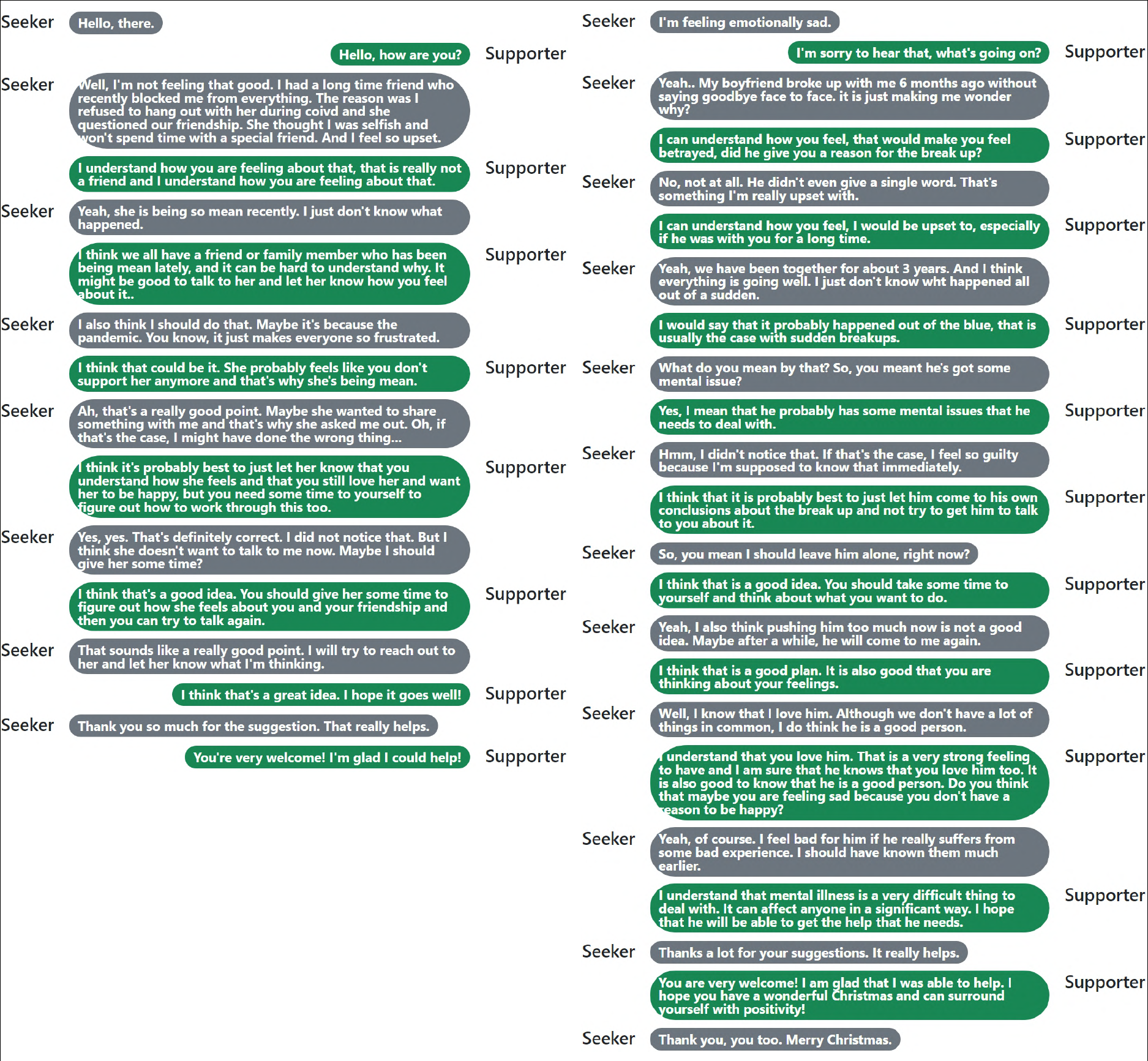}
        \caption{Sample conversations by BBMHR + \textit{ada} on the topics of ``problems with friends'' and ``break up with partner''.}
    \label{fig:sample_ada}
\end{figure*}

\begin{figure*}
    \centering
        \includegraphics[width=\textwidth]{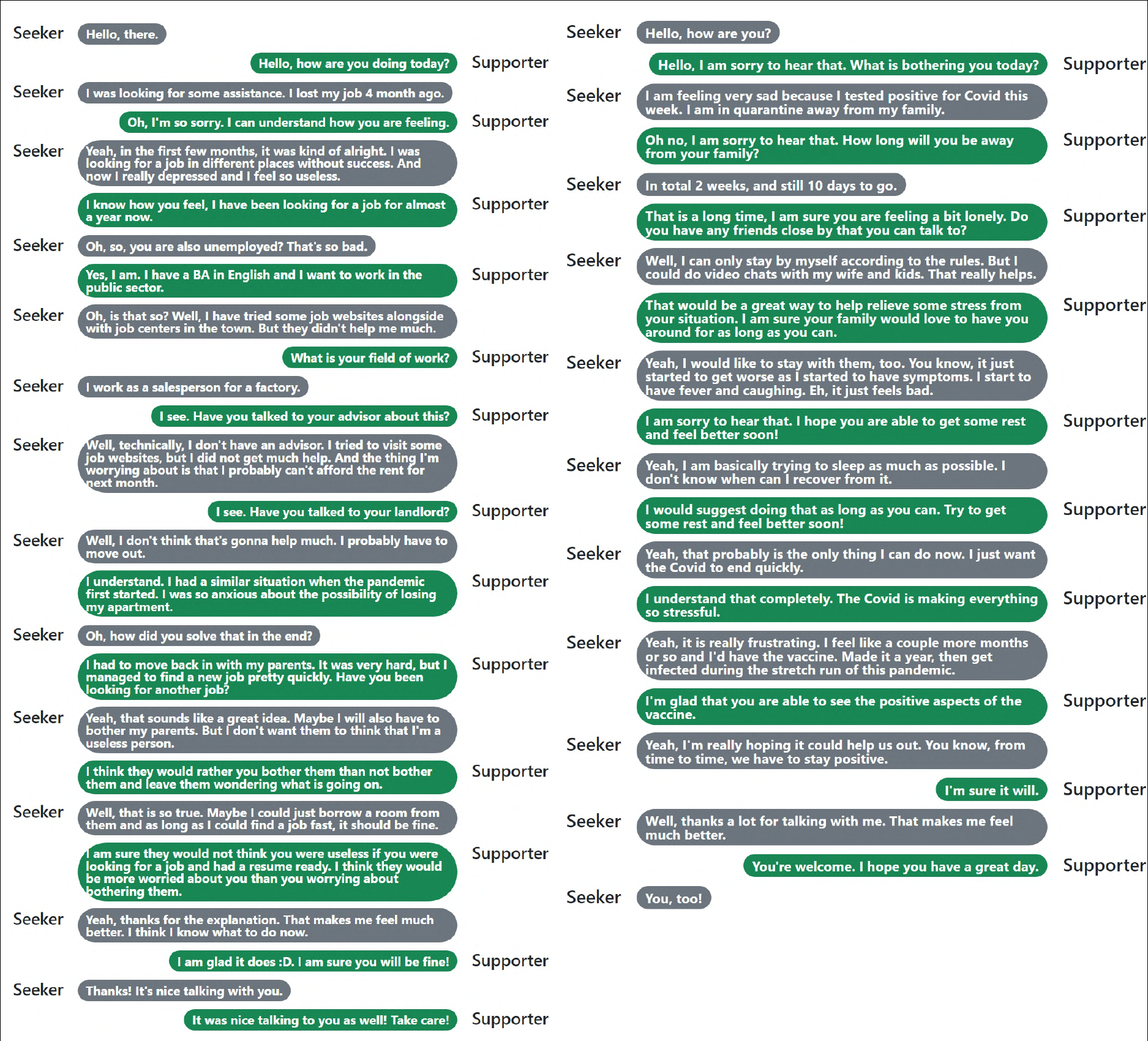}
        \caption{Sample conversations by BBMHR + \textit{gpt2} on the topics of ``job crisis'' and ``ongoing depression''.}
    \label{fig:sample_gpt_2}
\end{figure*}

\begin{figure*}
    \centering
        \includegraphics[width=\textwidth]{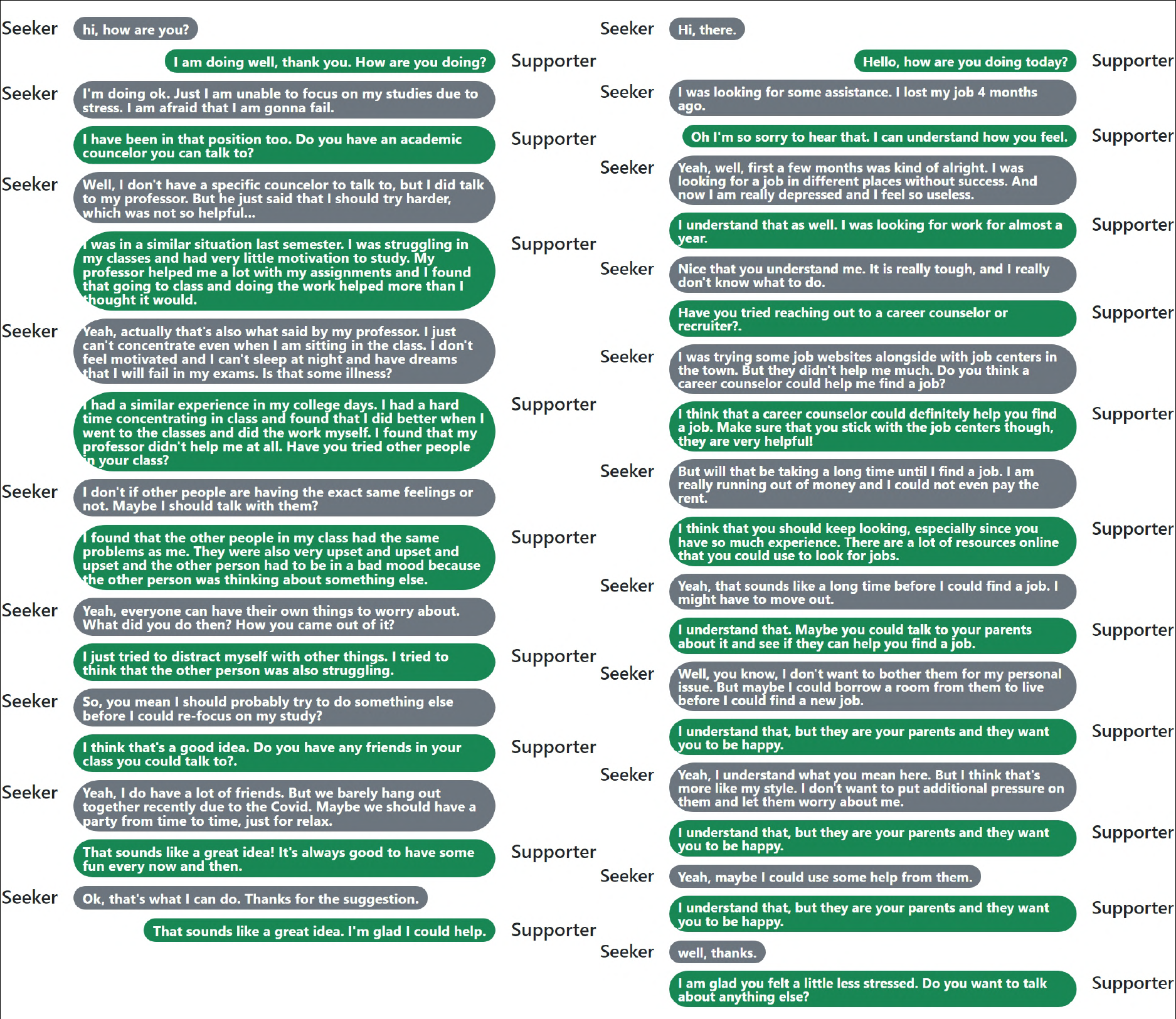}
        \caption{Sample conversations by BBMHR + \textit{gpt1} on the topics of ``academic'' and ``job crisis''.}
    \label{fig:sample_gpt_1}
\end{figure*}

\begin{figure*}
    \centering
        \includegraphics[width=\textwidth]{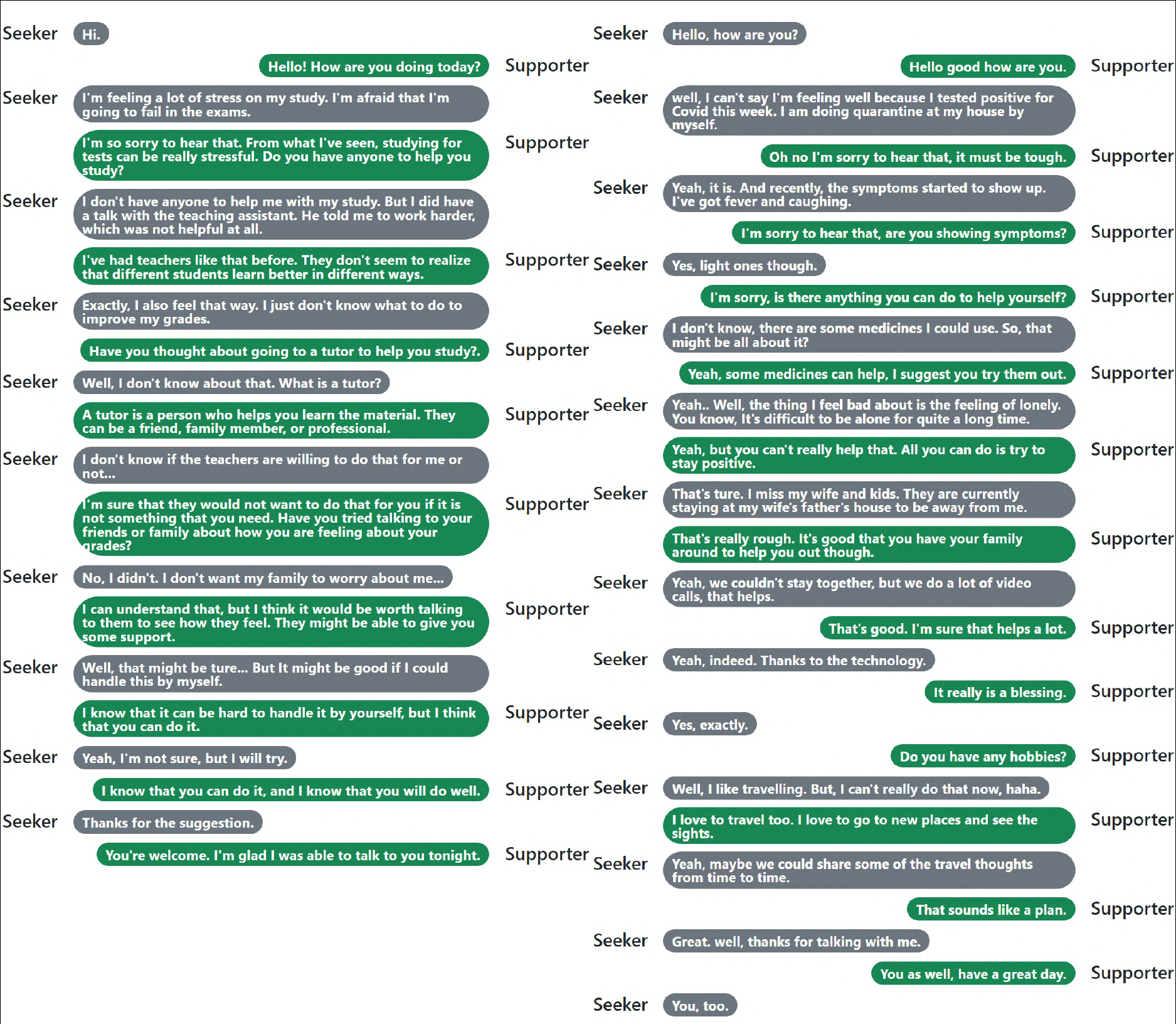}
        \caption{Sample conversations by BBMH on the topics of ``academic'' and ``ongoing depression''.}
    \label{fig:sample_bbmh}
\end{figure*}

\end{document}